\documentclass{article}
\usepackage{ijcai26}

\usepackage{times}
\usepackage{soul}
\usepackage{url}
\usepackage[hidelinks]{hyperref}
\usepackage[utf8]{inputenc}
\usepackage[small]{caption}
\usepackage{graphicx}
\usepackage{amsmath}
\usepackage{amsthm}
\usepackage{booktabs}
\usepackage{algorithm}
\usepackage{algorithmic}
\usepackage[switch]{lineno}
\usepackage{multirow}
\usepackage{enumitem}
\usepackage{amsmath}
\usepackage{amssymb}
\usepackage{xcolor}
\definecolor{mygreen}{RGB}{50,190,50}

\title{SRJudge: Empowering Large Language Models with Selective Reasoning for Fine-Grained Knowledge Concept Tagging}

\author{
Zhiwei Yang$^1$
\and
Jiahua Yang$^1$\and
Huiru Lin$^{2,3}$\And
Xing Chen$^{4,*}$  \And
Quanlong Guan$^{1,}$\thanks{Corresponding authors: Xing Chen, Quanlong Guan} \\
\affiliations
$^1$Guangdong Institute of Smart Education, Jinan University, Guangzhou, China\\
$^2$School of Physical Education, Jinan University, Guangzhou, China\\
$^3$Guangdong Provincial Key Laboratory of Speed Capability Research, Guangzhou, China\\
$^4$Sapient Intelligence Pte Ltd, Singapore\\
\emails
\{yangzw, linhuiru, gql\}@jnu.edu.cn, yangjiahua@stu2024 jnu.edu.cn, raincchio@gmail.com
}

\begin{document}

\maketitle

\begin{abstract}
 Knowledge concept tagging aims to assign specific concept or topic labels to educational content, which is essential for both educators and learners in traditional and online teaching practices. 
 Recent work has explored large language models (LLMs) for this task, achieving promising performance. However, LLMs still struggle to select the correct concept from a large-scale candidate set due to the high dimensionality of the decision space. In this paper, we propose a novel three-stage \textbf{S}elect-\textbf{R}eason-\textbf{Judge} (SRJudge) framework, which empowers LLMs with selective reasoning capability for fine-grained knowledge concept tagging.  
 Specifically, the Selector in Stage 1 first narrows the candidate concepts to a top-$K$ shortlist by fine-tuning a small language model (SLM), e.g., BERT, since the top-$K$ predictions hit the correct concept in most cases, thereby reducing the decision space of correct candidates. Next, the Stage 2 Reasoner employs a lightweight LLM for refined reasoning over the shortlisted candidates. It further integrates an improved reinforcement learning strategy with a dynamic task-specific reward function and a pruning mechanism to better align with human reasoning preferences. Finally, a larger LLM acts as a judger that evaluates the overall rationality of the reasoning process and its explanations to determine the final output. In addition, we construct two high-quality datasets for further validation, i.e., the biology dataset \textbf{S\_Bio} and the physics dataset \textbf{S\_Phy}.
 Experimental results demonstrate that our method consistently outperforms state-of-the-art baselines across benchmark datasets, verifying its effectiveness and superiority.  Resources are available at: https://github.com/Nicozwy/SRJudge. 
 
\end{abstract}

\begin{figure}[t]
\centering
\includegraphics[width=0.42\textwidth]{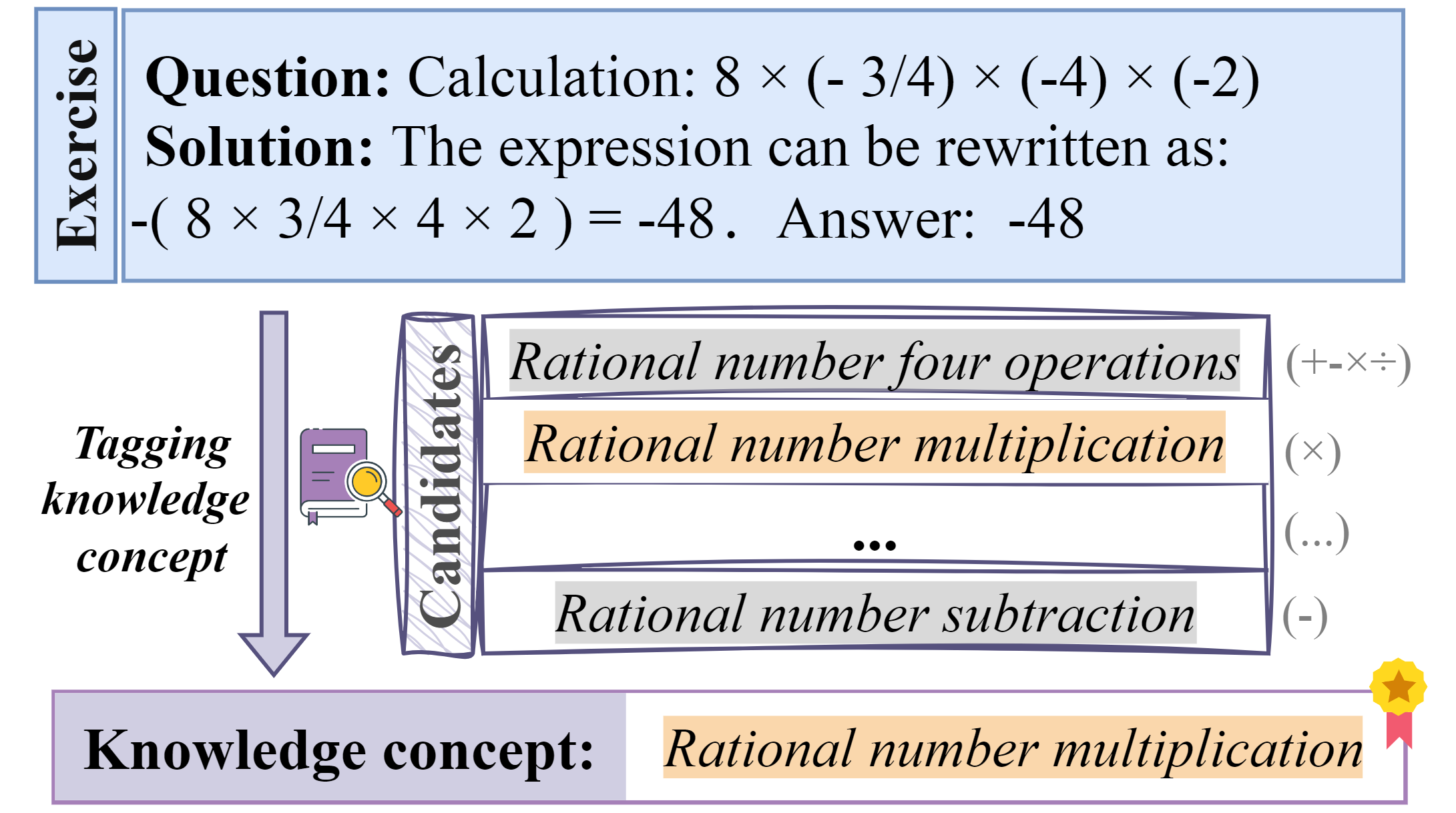} 
\caption{Illustration of the knowledge concept tagging task.}
\label{introduce}
\end{figure}

\section{Introduction}
Knowledge concepts form the foundation for key educational tasks on online platforms, such as MOOCs and intelligent tutoring systems. These tasks include knowledge tracing \cite{cui2024dgekt}, cognitive diagnosis \cite{wang2024boosting}, and intelligent recommendation \cite{ozyurt2025personalized}. 
To accurately index educational materials, it requires annotators to identify appropriate concepts based on the semantic and logical information within the question and its solution, as shown in Figure \ref{introduce}. Most existing knowledge concept tagging methods perform well yet struggle to select the correct concept among numerous similar candidates, making this a key performance bottleneck. 
Given the strong reasoning capabilities of large language models (LLMs), they could serve as enhanced tools for contextual understanding and disambiguation in knowledge concept tagging. However, LLMs often face challenges in selecting the correct concept from an extensive candidate set due to the high dimensionality of the decision space. 
Thus, the research on equipping LLMs with effective reasoning capabilities for navigating large-scale candidate concept sets is both challenging and essential. 

Existing tagging methods achieve decent performance utilizing support vector machines (SVM) \cite{saha2013discriminative}, recurrent neural networks (RNNs) \cite{sun2018automatic}, graph convolutional networks (GCNs) \cite{zhang2021question}, and RoBERTa \cite{ding2025tagging}. However, they are limited in capturing effective features and reasoning about similar knowledge. With the emergence of large language models (LLMs)  \cite{achiam2023gpt,bai2023qwen,touvron2023llama}, they have demonstrated impressive performance across various domains, such as mathematical reasoning \cite{xia2025evaluating}, code generation \cite{gu2023llm}, medicine \cite{thirunavukarasu2023large}, etc. Recently, \cite{ozyurt2024automated} leveraged LLMs' prior knowledge to generate step-by-step solutions for problems, then tagged each step with corresponding knowledge concepts. \cite{moore2024automated} designed two different prompting strategies to guide GPT-4 in generating knowledge concepts for multiple-choice questions, demonstrating LLMs can generate accurate knowledge concepts. To determine whether an existing knowledge concept aligns with a given question, \cite{li2025knowledge} reformulated the tagging process into sub-problems and utilized collaborative LLM-based agents to improve tagging consistency and robustness. In addition, \cite{yang2025lgcel} employed cascaded soft voting to integrate predictions from BERT-based models and LLMs, highlighting their complementary strengths. However, the large candidate set increases the dimensionality of the decision space, making it harder for the model to identify the relevant features that distinguish the correct concept.

As our preliminary studies shown in Table \ref{tab:motivation}, traditional SLMs can effectively capture the correct options within the top $K$ shortlist, but their accuracy remains limited when relying solely on the top-1 prediction. This arises from the semantic similarity among concepts within the top $K$ candidates. In contrast, while LLMs exhibit stronger generation and reasoning capabilities, a gap remains between model and human reasoning, particularly in selecting from large-scale candidates. Thus, these inherent limitations hinder existing methods from improving tagging performance.

\begin{table}[t]
\centering
\small
\setlength{\tabcolsep}{0.25pt}

\begin{tabular}{l|cc|cc|cc}
\toprule
\multirow{2}{*}{\textbf{Model}} & \multicolumn{2}{c|}{\textbf{S\_Math}} & \multicolumn{2}{c|}{\textbf{S\_Bio}} & \multicolumn{2}{c}{\textbf{S\_Phy}}\\
 & ACC & ACC@5 & ACC & ACC@5 & ACC & ACC@5 \\
\midrule
\textbf{Chinese-BERT} & 0.7200 & 0.9107 & 0.7216 & \textbf{0.9377} &0.6344  & \textbf{0.9154} \\
\textbf{Chinese-RoBERTa} & 0.7191 & 0.9148 & 0.7176 & 0.9247 &0.6352  &0.9100 \\
\textbf{Conan-embedding} & 0.7240 & \textbf{0.9156} & 0.7226 & 0.9296 &0.6367 &0.9051 \\
\bottomrule
\end{tabular}
\caption{Performance of BERT models with  \textit{supervised fine-tuning}, where ACC@5 denotes the accuracy of including the correct answer among the top 5 predictions.} 
\label{tab:motivation}
\end{table}

To this end, we propose a novel three-stage Select-Reason-Judge (SRJudge) framework that enables LLMs to effectively reason over large concept sets to achieve accurate tagging. 
Specifically, we first fine-tune a BERT-based Selector to select $K$ candidate concepts (Stage 1). Then, we apply pruning-based optimization
on the generalized reweighted policy optimization (GRPO) method to train a lightweight LLM Reasoner on the $K$ coarse-grained shortlist, and then design a task-specific reward function that introduces an additional dynamic position reward for the shortlist from Stage 1, reinforcing the Reasoner to recommend the most suitable concept along with its reasons (Stage 2). Finally, we incorporate an LLM Judger to evaluate the rationality of the recommendations and reasons from a global perspective and provide the final verdict (Stage 3).  Moreover, since no other validation datasets are available, we built two new datasets for method evaluation. Experimental results on datasets from three disciplines demonstrate that SRJudge consistently outperforms existing methods, validating the effectiveness of selective reasoning with reinforcement learning in enabling LLMs to tag the correct concept from a large candidate set. 

\textbf{Our contributions are summarized as follows:}
\begin{itemize}[leftmargin=*]
\sloppy
\item We present a novel three-stage framework, Select-Reason-Judge (SRJudge), that equips LLMs with selective reasoning capabilities through reinforcement learning for knowledge concept tagging amidst numerous candidates. 
\item We construct two new datasets, i.e., S\_Bio and S\_Phy, which will be made available to fill the gap in cross-validation of knowledge concept tagging methods across different disciplines. 
\item Experimental results show that SRJudge significantly outperforms leading baselines, highlighting its ability to enhance LLMs with selective reasoning to distinguish fine-grained knowledge concepts across different disciplines. 
\end{itemize}

\begin{figure*}[t]
\centering
\includegraphics[width=0.9\textwidth]{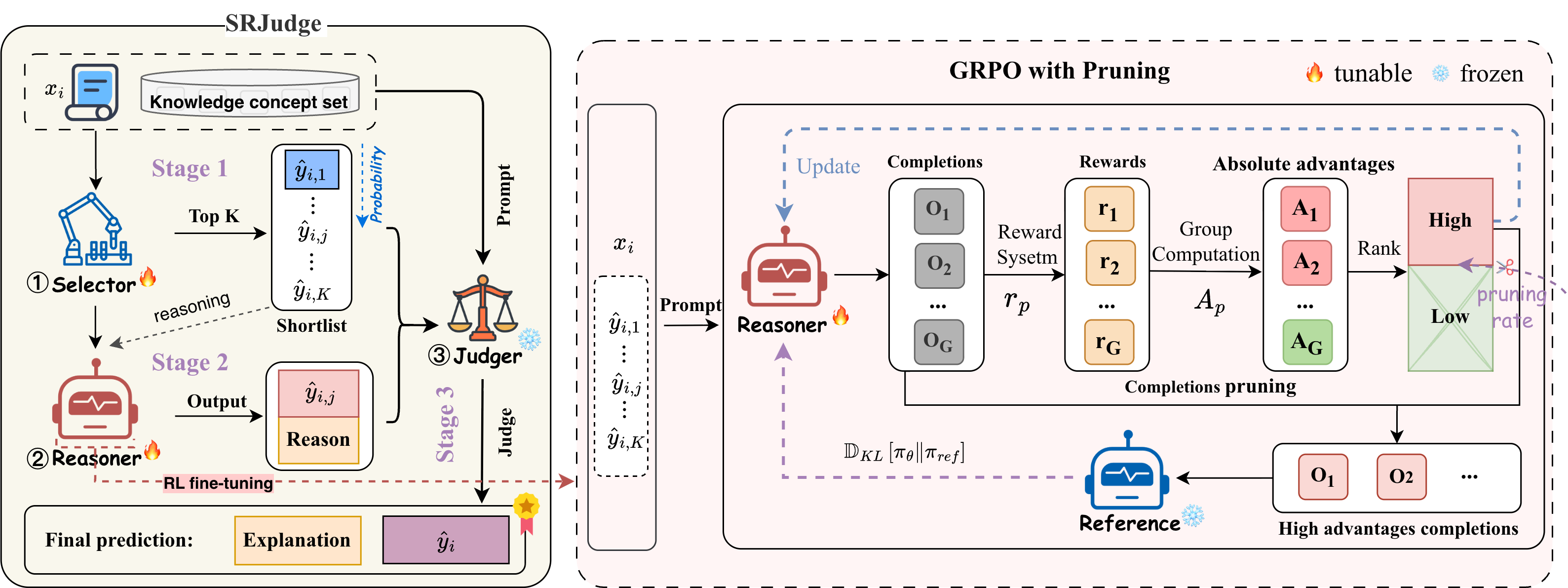} 
\caption{The proposed SRJudge framework for fine-grained knowledge concept tagging. RL denotes reinforcement learning, ``Reference'' denotes a frozen counterpart of the policy model (i.e., LLM Reasoner), $O_p$ is a binary pair defined as $O_p = (\hat{y}_{i,j}, \psi_{i,j}), 1\leq p \leq G$. 
}
\label{RLTAGG}
\end{figure*}

\section{Related Work}
\subsection{Traditional Knowledge Concept Tagging}
Traditional tagging methods primarily rely on traditional machine learning, lightweight deep learning, and small-scale pretrained models, involving feature engineering, contextual modeling, and graph structures \cite{li2022survey}.   
For example, \cite{saha2013discriminative} constructed a discriminative model based on SVM to perform tagging on questions from question-answering (QA) websites. Subsequently, deep learning methods were introduced into this domain. To enhance tagging performance, \cite{sun2018automatic} employed RNNs to capture contextual dependencies. To further address complex semantic and structural information, \cite{zhang2021question} proposed a graph-guided topic tagging model that integrates directed acyclic graphs and multi-dimensional attention to precisely tag topics in the QA dataset. Building upon this line of research, \cite{yang2022tag}  modeled sentences and concept graphs through graph-to-graph matching, mining logical relations and concept correspondences among terms using relational graph convolutional networks. \cite{liu2024improved} improved graph contrastive learning to obtain clearer and more discriminative class boundaries. \cite{huang2023pqsct} jointly modeled questions and answers using BERT and fused features to enhance the tagging performance.   
 
Although traditional methods have achieved successes in knowledge concept tagging, 
they still suffer from limited capabilities in capturing effective features and reasoning about similar knowledge, hindering their effectiveness in fine-grained knowledge concept tagging scenarios. 

\subsection{LLM-based Tagging}
Recently, LLM-based tagging methods have explored the robust generative and reasoning capabilities of LLMs to enhance tagging accuracy and interpretability through prompt engineering \cite{moore2024automated}, CoT reasoning \cite{ozyurt2024automated}, and multi-agent collaboration \cite{li2025knowledge}.  For example, \cite{ozyurt2024automated} improved the performance of knowledge tracing models by utilizing LLMs first to solve exercises via CoT, followed by automatic tagging of knowledge concepts with these exercises, thereby strengthening the semantic linkage between exercises and knowledge concepts. Building on this, \cite{moore2024automated} utilized two different prompting strategies to guide GPT-4 to generate a knowledge concept for multiple-choice questions and compares them with the original knowledge concepts.  Results revealed a human preference for LLM-generated concepts over original annotations, highlighting the superior generative quality of LLMs in this context. To address the complexity and improve the reliability of exercise tagging, \cite{li2025knowledge} reformulated the exercise tagging judgment process as a collaboration among multiple LLM-based agents tackling independent subproblems. 
Prior work indicates that collaborative model ensemble strategies improve text classification performance. Specifically, \cite{zhang2025pushing} reported substantial gains from iteratively ensembling seven strong LLMs. Similarly, \cite{yang2025lgcel} reported that integrating fine-tuned BERT-series models with LLMs through local and global ensemble voting significantly improves labeling performance.
Although fine-tuned LLMs achieve decent performance, they may not cover all potential user needs with static datasets, suffering from high computational costs \cite{ghosh2024closer}. Additionally, prompt-based reasoning struggles to align with human reasoning preferences.

\section{Proposed Method}
\subsection{Problem Statement}
Given an exercise $x_i \in \{x_1, x_2, \dots, x_m\}$ associated with a knowledge concept label \( y_i \in \Phi= \{y_i\}_{i=1}^{|\Phi|} \), the task of knowledge concept tagging aims to predict the label $\hat{y}_i$ for $x_i$, i.e., $f(x_i) \longrightarrow \hat{y}_i$, where $|\Phi|$ represents the category number of candidate knowledge concepts.

\textbf{Overview}: As illustrated in Figure \ref{RLTAGG}, our proposed SRJudge comprises three key components: Selector, Reasoner, and Judger. For the exercise $x_i$, the Selector first picks $K  (\ll |\Phi|) $ candidates, $\{\hat{y}_{i,1}, \dots, \hat{y}_{i,j},\dots, \hat{y}_{i,K}\}$, reducing the decision space. The Reasoner then performs reasoning over the shortlist, producing both the recommended concept $\hat{y}_{i,j}$ and corresponding reasons. Finally, the Judger assesses the overall rationality of the reasoning process and its explanations to determine the final output $\hat{y}_i$.

\subsection{Selector for Scope Reduction}
To reduce the candidate concept space, we use a lightweight pre-trained model (e.g., BERT) to select the top $K$ candidates from a large pool in Stage 1, thereby improving exploration efficiency while maximizing retention of correct concepts in the shortlist. In this work, we employ unsupervised pre-training and supervised fine-tuning to achieve a more effective Selector. Specifically, masked language modeling is employed in the continued pre-training phase to enhance the model's semantic understanding. We fine-tune the Selector on a labeled dataset to improve its accuracy in determining the relevance of knowledge concepts to a given query. Thus, these $K$ candidates are ranked and selected based on predicted probabilities using softmax normalization as follows:   
\begin{equation}
\mathcal{C}_{\text{top }K}
 = \left\{ \hat{y}_{i,j} \in \Phi \;\middle|\; \operatorname{rank}_{\downarrow}(p_{i,j}) \leq K \right\} 
\end{equation}
where $\operatorname{rank}_{\downarrow} ()$ indicates ranking candidates in descending order, $p_{i,j}$ is the probability of candidate $\hat{y}_{i,j}$. This selected candidate set $\mathcal{C}_{\text{top }K}$ is then fed into Stage 2 to ensure that higher-quality candidates are concentrated within a smaller scope, facilitating subsequent reinforcement learning (RL).

\subsection{Reasoner for Concept Recommendation}

In Stage 2, we employ an RL algorithm — Group Relative Policy Optimization(GRPO) ~\cite{shao2024deepseekmath} to enhance the reasoning accuracy of LLMs on the $K$ candidates for fine-grained knowledge concept tagging. That is, the candidate concept set $\{\hat{y}_{i,1}, \dots, \hat{y}_{i,j},\dots, \hat{y}_{i,K}\}$ provided by the Selector is treated as the decision space. For training efficiency, we utilize a lightweight LLM (e.g., Qwen2.5-1.5B-Instruct) as the policy model and apply GRPO to guide accurate concept tagging with corresponding justifications, employing a newly designed task-specific reward function. The goal is to select the most suitable concept label $\hat{y}_{i,j}$ for each exercise and provide the corresponding reasons. Specifically, for each exercise $x$ ($x\sim P(Q)$, omit the subscript $i$ for convenience), GRPO samples $G$ output completions $ \{o_1, o_2, \dots,o_p,\dots, o_G\}$ from the old policy $\pi_{\theta_{\text{old}}}$,
where $o_p$ is defined as $o_p = (\hat{y}_{i,j}, \psi_{i,j})$, where $\hat{y}_{i,j}$ and $\psi_{i,j}$ denote the predicted knowledge concept and reasons in the $p$-th output, respectively.
Then the policy model is optimized by maximizing the following objective:
\begin{align}
\mathcal{J}_{\text{GRPO}}(\theta) &= \mathbb{E}_{x \sim P(Q), \{o_p\}_{p=1}^G \sim \pi_{\theta_{\text{old}}}(\cdot|x)} \Bigg\{
\frac{1}{G} \sum_{p=1}^G \frac{1}{|o_p|} \sum_{t=1}^{|o_p|} \nonumber \\
& \hspace{-1pt} \bigg\{ \min \bigg[
\frac{\pi_{\theta}(o_{p,t}|x, o_{p,<t})}{\pi_{\theta_{\text{old}}}(o_{p,t}|x, o_{p,<t})} \hat{A}_{p,t}, \nonumber \\
& \hspace{-1pt} \mathrm{clip}( 
\frac{\pi_{\theta}(o_{p,t}|x, o_{p,<t})}{\pi_{\theta_{\text{old}}}(o_{p,t}|x, o_{p,<t})}, 
1-\varepsilon, 
1+\varepsilon
) \hat{A}_{p,t}
\bigg] \nonumber \\
& \hspace{-1pt} - \beta \mathbb{D}_{KL} \left[\pi_{\theta} \| \pi_{ref}\right] \bigg\}
\Bigg\}
\label{grpo}
\end{align}

where \(\theta\) denotes the parameters of the current policy model \(\pi_{\theta}\), \(\pi_{\text{ref}}\) is the reference policy model, $\varepsilon$ is a clipping-related hyper-parameter introduced in GRPO for stabilizing training, and $\beta$ is the coefficient of the KL penalty.
\(\hat{A}_{p,t}\) is the advantage value computed based solely on the relative rewards within each group of outputs.
GRPO regularizes the policy update by directly adding the Kullback--Leibler (KL) \cite{kullback1951information} divergence between the training policy and the reference policy to the loss function. 
 $A_{p}$ is the advantage computed using a group of rewards $\{r_{1}, r_{2}, \ldots, r_p,\ldots, r_{G}\}$ corresponding to the completions within each group:
\begin{align}
A_{p} &= \frac{r_{p} - \operatorname*{mean}(\{r_{1},r_{2},\ldots,r_{G}\})}{\operatorname*{std}(\{r_{1},r_{2},\ldots,r_{G}\})}
\label{eq:advantage}
\end{align}

\textbf{Task-specific Reward. } 
To leverage GRPO for enhancing LLM reasoning on fine-grained concepts, the reward function is specifically designed to optimize task performance by incorporating three types of rewards: output format, answer accuracy, and correct answer position, as follows:
\begin{equation}
r_{p} =  R_{\text{format}}(o_p) +  R_{\text{acc}}(o_p) + R_{\text{position}}(o_p)
\label{eq:reward_function} 
\end{equation}

\(R_{\text{format}}(o_p)\) denotes the format reward to enforce a standardized output format, requiring the final answer to be enclosed within \verb|\boxed{}| (+1) for answer extraction. The model's reasoning process must be clearly marked using the tags \verb|<reason>| and \verb|</reason>|  (+0.5) to ensure a well-structured explanation. \(R_{\text{acc}}(o_p)\) represents the accuracy reward derived from the correctness of the model's answer, providing positive feedback when the answer is correct (+3) and no reward otherwise. 
Additionally, \(R_{\text{position}}(o_p)\) denotes the position reward, 
which is dynamically \textbf{conditioned on the model predicting correctly and and the correct answer being ranked at position $r \in [1, K-1]$ in the candidate shortlist}. This is formulated as follows:
\begin{equation}
\begin{aligned}
R_{\text{position}}(o_p)
&= \alpha \cdot 
\frac{\frac{1}{p_r}}{\sum_{j=0}^{K-1} \frac{1}{p_j}}
\cdot 
\frac{\log_2(r+1)}{\log_2 K}, \\[4pt]
\alpha &= \frac{10 \times \text{S}_{\text{c}}}{\text{S}_{\text{T}} \times \left(1-\text{pruning rate}\right)} + 1
\end{aligned}
\label{eq:rpos_and_alpha}
\end{equation}

where $r$ is the position of the correct answer in the candidate shortlist (starting from 0), $K$ is the length of the shortlist, $p_r$ is the probability associated with rank $r$, and $\alpha$ is a dynamic scaling factor determined by the current training step and model pruning rate. Specifically, $\text{S}_{\text{c}}$ and $\text{S}_{\text{T}}$ represent the current training step and the total number of training steps, respectively. 
The proposed position-based reward is designed to ensure bounded and stable policy gradients. 
The normalization of $R_{\text{position}}(o_p)$ within a manageable range prevents excessively large gradients during early training, which effectively improves convergence stability. 
A small $\alpha$ at the beginning helps reduce gradient variance and stabilize learning, while a larger $\alpha$ in the later phase enhances reward diversity and prevents premature convergence. 
Since candidates are ranked by confidence, their positions carry meaningful prior information. 
To encourage moderate exploration, the additional position reward is granted only when the correct answer appears between the 2nd and $K$-th positions ($r \in [1, K-1]$). 
When the correct answer is at the first position ($r = 0$), the logarithmic term $\log_2(r+1)$ evaluates to zero, and thus $R_{\text{position}} = 0$.
This design maintains training stability in the early stages while gradually encouraging exploration of lower-ranked candidates as training progresses. Thus, it balances exploration and exploitation, stabilizing policy gradient updates throughout the learning process.

\textbf{Pruning Mechanism. } 
When applying GRPO to knowledge concept tagging, the agent can be distracted by low-quality completions with positive relative advantages. During RL training, such low-quality outputs typically manifest as ``format errors with correct answers’’ or ``correct formats with incorrect answers’’ in the early stages, and predominantly as ``correct formats with incorrect answers’’ in later stages. These completions introduce noise that hinders effective agent–environment interaction, and ultimately degrades the stability and efficiency of policy optimization.
Therefore, we introduce a pruning mechanism in reinforcement learning, which computes the absolute advantage of each completion, ranks them, and dynamically filters out those with low absolute advantage. To facilitate model learning, we further impose the confidence constraint on the GRPO optimization objective in Equation~\eqref{grpo}, as follows:
\begin{align}
|\hat{A}_{p,t}| \ge \gamma
\label{eq:grpo_pruning}
\end{align}
where $\gamma$ is selected from the set of absolute advantages $\{|\hat{A}_{p,t}|\}$ according to the predefined pruning rate. 
Specifically, $\gamma$ represents the minimum absolute advantage that a completion among the $G$ generated outputs must satisfy to be retained for gradient updates. 
This ensures that only completions with strong training signals are preserved for policy updates, thereby improving training efficiency and mitigating interference from ineffective or low-quality samples. 

\subsection{Judger for Overall Assessment}
In Stage 3, we utilize a frozen LLM (e.g., Qwen3-32B) as the Judger to comprehensively evaluate and select the correct knowledge concept and generate explanations after reinforced reasoning, which further enhances the accuracy and robustness of the final decision from a global perspective. 
Specifically, the Judger takes as input the top 1 prediction from the Selector, the selected $K$ candidate concepts, the recommended concepts, and the corresponding reasons generated by the Reasoner. 
Thus, the Judger makes final decisions from global perspectives utilizing the following prompts:  

\textbf{\textit{System-prompt:}}You are an assistant for tagging knowledge concepts in  mathematics exercises and do not need to solve the problem. In the following dialogue, I would like you to reason step by step. Please refer to both the predictions and reasoning from the two models, as well as the list of knowledge concepts, to help me select the knowledge concept that best fits the question. Please enclose the selected knowledge concept within boxed \{$\hat{y}_i$\}. 

\textbf{\textit{User-prompt:``}} The  mathematics question and its solution are: [from the dataset] 

\textbf{\textit{The list of knowledge concepts is:}} 
[from Stage 1] 

\textbf{\textit{Model 1 predicted\_label:}} 
[from Stage 1] 

\textbf{\textit{Model 2 predicted\_label:}} 
[from Stage 2]

\textbf{\textit{The reason for Model 2’s choice is:}} 
[from Stage 2] \textbf{\textit{"}}


\section{Experiments}
\subsection{Datasets and Baseline Methods}
To verify the effectiveness of SRJudge, we conduct experiments on three educational datasets: \textbf{S\_Math} \cite{yang2025lgcel}, \textbf{S\_Bio}, and \textbf{S\_Phy},   
For cross-validation, we constructed the new biological dataset, S\_Bio, and the new physics dataset, S\_Phy. Both datasets were developed in a similar manner, incorporating questions, solutions, difficulty scores, and knowledge concepts. For unsupervised training, there are 23,596, 15,669, and 32,138 unlabeled instances in U\_Math, U\_Bio, and U\_Phy, respectively (See our repository for more details). 
We perform an 8:1:1 stratified split into training, validation, and test sets for each dataset, ensuring that the distribution of concept categories is balanced across each subset. Table~\ref{tab:dataset} provides the dataset statistics.

\textbf{Baseline Methods:} We compare our proposed framework with the state-of-the-art baselines as follows: 

\begin{itemize}[leftmargin=*]
\item SLM-based methods: We respectively compare the use of BERT \cite{devlin2019bert}, RoBERTa \cite{liu2019roberta}, and Conan-Embedding \cite{li2024conan} as backbone encoders, each of which is followed by a classification head and trained via supervised fine-tuning ($SFT$).

\item LLM-based methods: We employ mainstream LLMs, including Qwen  \cite{bai2023qwen}, LLaMA  \cite{touvron2023llama}, DeepSeek (DS) \cite{guo2025deepseek}, and ChatGPT \cite{achiam2023gpt}.
For LLMs with fewer than 8B parameters, we explore $SFT$ and $GRPO$ strategies, respectively. Due to computational cost, 
we adopt $SFT$ for open-source models and zero-shot CoT prompting ($COT$) for closed-source models for LLMs with over 8B parameters. 
\item Additional Baselines: 
Among existing studies on exercise knowledge concept tagging, 
PQSCT~\cite{huang2023pqsct}, LGCEL~\cite{yang2025lgcel}, and LHABS~\cite{DING2025126232} are specifically designed for this task.
PQSCT jointly models questions and answers using BERT with feature fusion techniques, 
LGCEL leverages multiple heterogeneous expert models to enhance tagging accuracy through model complementarity, 
while LHABS improves RoBERTa with extra attention and label smoothing.
We further compare two latest classification baselines: 
GIFT~\cite{liu2024improved} employs a  graph neural network combined with contrastive learning,  
and RGPT~\cite{zhang2025pushing} integrates seven LLaMA models through a cascaded guidance mechanism for text classification.

\end{itemize}

\begin{table}[t]
\centering
\setlength{\tabcolsep}{2.5pt}
\small
\renewcommand{\arraystretch}{0.85}
\begin{tabular}{lrrrrcr} 
\toprule
\textbf{Dataset} & \textbf{Sum} & \textbf{Train} & \textbf{Valid} & \textbf{Test} & \textbf{Category} & \textbf{Level} \\
\midrule
S\_Math &12,205 & 9,763 & 1,221 & 1,221  & 155 & G7--G9\\
S\_Bio &9,941 & 7,952 & 994 & 995  & 72 & G10--G12\\
S\_Phy &18,440 & 14,752 & 1,844 & 1,844  & 135 & G10--G12\\
\bottomrule
\end{tabular}
\caption{Dataset statistics.} 
\label{tab:dataset} 
\end{table}

\begin{table*}[t]
\centering
\small
\setlength{\tabcolsep}{3.1pt}
\renewcommand{\arraystretch}{0.9}
\begin{tabular}{l|c|cccc|cccc|cccc}
\toprule
\multirow{2}{*}{\textbf{Model}}  & \multirow{2}{*}{\textbf{Strategy}} & \multicolumn{4}{c|}{\textbf{S\_Math}} & \multicolumn{4}{c|}{\textbf{S\_Bio}} & \multicolumn{4}{c}{\textbf{S\_Phy}} \\
&  & ACC & P & R & F1 & ACC & P & R & F1 & ACC & P & R & F1\\
\midrule

Chinese-RoBERTa  & $SFT$ & 0.7191 & 0.7213 & 0.7062 & 0.6992 & 0.7176 & 0.6889 & 0.6624 & 0.6613 &0.6352	&0.6378	&0.6232	&0.6227 \\
Chinese-BERT      & $SFT$ & 0.7200 & 0.7213 & 0.6938 & 0.6911 & 0.7216 & 0.6900 & 0.6743 & 0.6682 &0.6344	&0.6359	&0.6217	&0.6203 \\
Conan-embedding         & $SFT$ & 0.7240 & 0.7164 & 0.7118 & 0.7017 & 0.7226 & 0.7003 & 0.6725 & 0.6786 &0.6367	&0.6267	&0.6177	&0.6165 \\

\midrule

\multirow{2}{*}{Qwen2.5-7B-Instruct}  & $SFT$ & 0.7428 & 0.7273 & 0.7165 & 0.7041 & 0.7286 & 0.6989 & 0.6808 & 0.6790 &0.6524	&0.6555	&0.6340	&0.6336 \\
 & $GRPO$ & \underline{0.7117} & \underline{0.6989} & \underline{0.6833} & \underline{0.6600} & \underline{0.6613} & \underline{0.5874} & \underline{0.5923} & \underline{0.5674} &\underline{0.6410} &\underline{0.5721} &\underline{0.5598} &\underline{0.5388}\\
\midrule

\multirow{2}{*}{LLaMA-3.1-8B-Instruct}  & $SFT$ & 0.7371 & 0.7223 & 0.7185 & 0.7079 & 0.7266 & 0.6967 & \underline{0.6775} & \underline{0.6794} &0.6534	&0.6537	&0.6276	&0.6312\\

 & $GRPO$ & 0.6617 & 0.6551 & 0.6530 & 0.6278 & 0.6119 & 0.5601 & 0.5719 & 0.5398 &0.6133 &0.5523 &0.5331 &0.5250 \\
\midrule
\multirow{2}{*}{DS-R1-Qwen-7B}  & $SFT$ & 0.7428 & 0.7415 & 0.7192 & 0.7141 & 0.7226 & 0.7092 & 0.6748 & 0.6739 &0.6504	&0.6470	&0.6272	&0.6266\\

  & $GRPO$ & 0.6642 & 0.6108 & 0.6132 & 0.5917 & 0.6442 & 0.5482 & 0.5772 & 0.5410 &0.6318 &0.5706 &0.5584 &0.5345\\
\midrule
\multirow{1}{*}{DS-R1-Qwen-32B}  & $SFT$ &0.7387&0.7218&0.7193&0.7175 &0.7256&0.7024&0.6797&0.6752 &0.6947&0.6577&0.6419&0.6390\\
\multirow{1}{*}{DS-R1-LLaMA-72B}  & $SFT$ &0.7477&0.7305&\underline{0.7248}&\underline{0.7264} &0.7367&0.7081&0.6796&0.6821 &\underline{0.6795}&\underline{0.6742}&0.6558&\underline{0.6586}\\
\multirow{1}{*}{Qwen3-14B}  & $SFT$ &0.7461	&0.7249	&0.7200	&0.7156		&0.7387	&0.7018	&0.6847	&0.6818		&0.6676	&0.6554	&0.6418	&0.6366\\
\multirow{1}{*}{Qwen3-32B}  & $SFT$ &0.7494	&0.7456	&0.7175	&0.7187		&0.7387	&0.7069	&0.6828	&0.6841		&0.6714	&0.6708	&0.6449	&0.6447\\
\multirow{1}{*}{Qwen2.5-72B-Instruct}  & $SFT$ &\underline{0.7518}	&\underline{0.7517}	&0.7188	&0.7204		&\underline{0.7396}	&\underline{0.7127}	&\underline{0.7013}	&\underline{0.6855}		&0.6768	&0.6709	&\underline{0.6585}	&0.6526\\
\midrule

DS-R1-0528   & $COT$ & \underline{0.6355} & \underline{0.6458} & \underline{0.6239} & \underline{0.5954} & \underline{0.6391} & \underline{0.5876} & \underline{0.5981} & \underline{0.5747}  &\underline{0.4506}	&\underline{0.4596}	&\underline{0.4206}	&\underline{0.3923} \\
GPT-4.1  & $COT$ & 0.5659 & 0.5942 & 0.5350 & 0.5047 & 0.5728 & 0.5311 & 0.5288 & 0.5068 &0.4192	&0.4115	&0.4040	&0.3664 \\
\midrule
PQSCT  & – &0.7174 &0.7182 &0.6937 &0.6896 &0.7296 &0.6935 &0.6716 &0.6667 &0.6676 &0.6525 &0.6245 &0.6244 \\
GIFT  & – & 0.7027 & 0.6986 & 0.6824 & 0.6749 & 0.7256 & 0.6939 & 0.6821 & 0.6758  &0.6649 &0.6641 &0.6285 &0.6210 \\
LHABS  & – & 0.7158 & 0.7091 & 0.6969 & 0.6864 & 0.7327 & 0.6939 & 0.6852 & 0.6763 &0.6708 &0.6543 &0.6251 &0.6229 \\

RGPT  & – & 0.7412 & 0.7215 & 0.7211 & 0.7086 & 0.7236 & 0.6892 & 0.6833 & 0.6712 &0.6703 &0.6509 &0.6353 &0.6287 \\

LGCEL  & – & 0.7660 & 0.7602 & 0.7379 & 0.7311 & 0.7417 & 0.7042 & 0.6831 & 0.6840 &0.6725	&0.6700	&0.6489	&0.6476 \\
SRJudge (Ours) & – & \textbf{0.7748} & \textbf{0.7772} & \textbf{0.7721} & \textbf{0.7602} & \textbf{0.7437} & \textbf{0.7285} & \textbf{0.7029} & \textbf{0.6987} &\textbf{0.6920}	&\textbf{0.6791}	&\textbf{0.6702}	&\textbf{0.6643} \\

\bottomrule
\end{tabular}
\caption{Performance comparison on the S\_Math, S\_Bio, and S\_Phy datasets. The bold numbers denote the best results, and the underlined numbers denote the best performance of baselines using different strategies (statistically significant at $p < 0.05$).}
\label{tab:math-bio-results}
\end{table*}

\textbf{Experimental Settings:} 
In supervised fine-tuning, we set the batch size, number of epochs, and learning rate to 16, 15, and 1e-4, respectively. Evaluation is performed every 100 steps, with early stopping applied (patience = 10). For GRPO, the batch size, number of epochs, number of generations, and pruning rate are set to 12, 3, 12, and 0.5, respectively. 
Additionally, the clipping parameter $\varepsilon$, the KL penalty coefficient $\beta$, and the learning rate are set to 0.2, 0.04, and $1\times10^{-6}$, respectively.
The hyperparameter $K$ is set to 5. 
Our proposed method is implemented and evaluated on a cluster of four NVIDIA L20 servers, using accuracy (\textbf{ACC}), precision (\textbf{P}), recall (\textbf{R}), and macro F1 score (\textbf{F1}) as evaluation metrics.

\subsection{Overall Results and Analysis}

Table \ref{tab:math-bio-results} shows the experimental results of SRJudge against various baselines on S\_Math, S\_Bio, and S\_Phy. Overall, SRJudge outperforms LGCEL, achieving an F1 score of 0.7602 on S\_Math, 0.6987 on S\_Bio, and 0.6643 on S\_Phy, consistently yielding the best results. 
This demonstrates the effectiveness of the proposed three-stage framework, SRJudge, which reduces the decision space through supervised fine-tuning, reinforces LLM reasoning via tailored GRPO, and provides an overall assessment from a global perspective.   

Specifically, the latest baselines, including PQSCT, GIFT, LHABS, and RGPT, demonstrate relatively unsatisfactory performance compared to LGCEL, which remains the leading baseline, achieving F1 scores of 0.7311, 0.6840, and 0.6476 on the S\_Math, S\_Bio, and S\_Phy datasets, respectively. This is because PQSCT fuses separate features of questions and answers without exploring more effective information, while GIFT may introduce noise by depending on external knowledge graphs for augmented view generation. 
Furthermore, LHABS shows reduced performance, likely due to label smoothing, which diminishes subtle distinctions among knowledge concepts, thereby complicating the model's ability to differentiate between them. Although RGPT shares conceptual similarities with LGCEL, it suffers from limited model diversity. In contrast, LGCEL leverages heterogeneous model predictions that complement one another, thereby enhancing overall performance.

Moreover, we conduct a comprehensive comparison of more SLMs and LLMs for in-depth analysis. 
SLM-based tagging methods using the $SFT$ strategy achieve lower performance compared to LLM-based methods, but due to their smaller parameter size and lower training cost, they remain highly practical. 
For LLMs employing the $SFT$ strategy, Qwen2.5-72B-Instruct outperforms Qwen2.5-7B-Instruct, suggesting that increasing model size can enhance their logical reasoning and tagging performance to some extent. 
In the $COT$ strategy setting, even state-of-the-art models such as DeepSeek-R1-0528 and GPT-4.1 struggle to achieve satisfactory results due to the large decision space and the lack of supervised fine-tuning.
While LLMs can benefit from $GRPO$ in knowledge concept tagging, their performance remains significantly inferior to that achieved with the SFT strategy, obtaining F1 scores of only 0.6600, 0.5674, and 0.5388 on the S\_Math, S\_Bio, and S\_Phy datasets, respectively. This is also due to GRPO's extensive exploration of the search space for selecting the correct concept labels, leading to sparse rewards and challenges in convergence. 

In summary, SRJudge consistently outperforms alternatives on all datasets, underscoring its effectiveness in enhancing LLMs with selective reasoning through reinforcement learning. Consequently, this facilitates assigning the correct knowledge concept from numerous fine-grained candidates.  

\begin{table*}[t]
\centering
\small
\setlength{\tabcolsep}{2.5pt}
\renewcommand{\arraystretch}{0.75}
\begin{tabular}{l|cccc|cccc|cccc}

\toprule
  \multirow{2}{*}{\textbf{Model}}& \multicolumn{4}{c|}{\textbf{S\_Math}} & \multicolumn{4}{c|}{\textbf{S\_Bio}}& \multicolumn{4}{c}{\textbf{S\_Phy}} \\
&  ACC & P & R & F1 & ACC & P & R & F1 & ACC & P & R & F1\\
\midrule

 Chinese-BERT (Selector, \textbf{S})  & 0.7412 & 0.7503 & 0.7386 & 0.7289 & 0.7337 & 0.7045 & 0.6865 & 0.6829 &0.6752	&0.6748	&0.6519	&0.6437\\

\midrule

 Qwen2.5-1.5B-Instruct (Reasoner, \textbf{R} ) & 0.6454& 0.6045 &0.6057& 0.5785 & 0.5879 & 0.4704 & 0.5468 & 0.4952 &0.5548 &0.4930 &0.4705 &0.4539\\

\midrule

 Qwen3-14B (Judger, \textbf{J$_1$} )  &0.5913 &0.6161 &0.5694  &0.5488 &0.5508 &0.5182 &0.5099  &0.4883 &0.3791 &0.3831 &0.3565 &0.3373 \\
 Qwen3-32B (Judger, \textbf{J$_2$})   &0.6093 &0.6179 &0.5913  &0.5660 &0.5688 &0.5210 &0.5251 &0.4940 &0.4018 &0.4181 &0.3952 &0.3705 \\
\midrule

 \textbf{S + R}
 & 0.7649	&0.7738	&0.7573	&0.7491 & 0.7417	&0.7128	&0.6986	&0.6925 &0.6871	&0.6779	&0.6667	&0.6629\\

\midrule


 \textbf{S + R}  + \textbf{J$_1$}  &0.7739	&0.7760	&0.7719	&0.7592		&0.7427	&0.7146	&0.7003	&0.6946 &0.6887	&0.6779	&0.6681	&0.6634 \\
 \textbf{S + R}  + \textbf{J$_2$}  &\textbf{0.7748}	&\textbf{0.7772}	&\textbf{0.7721}	&\textbf{0.7602}	&\textbf{0.7437}	&\textbf{0.7285}	&\textbf{0.7029}	&\textbf{0.6987} &\textbf{0.6920}	&\textbf{0.6791}	&\textbf{0.6702}	&\textbf{0.6643} \\

\bottomrule
\end{tabular}
\caption{Ablation Study of SRJudge. We provide additional ablation results of SRJudge variants using different Judgers.} 
\label{tab:stage1+2+3 1+3} 
\end{table*}

\subsection{Ablation Study}
Table \ref{tab:stage1+2+3 1+3} presents the results of our ablation studies to verify the effectiveness of the components and their various combinations in SRJudge. 
Overall, SRJudge consistently achieves the best performance across all ablation counterparts, demonstrating that each component contributes to the effectiveness of the proposed framework. Specifically, the Selector of SRJudge outperforms BERT-series models, as shown at the top of Table 
 \ref{tab:math-bio-results}, because continued pretraining can further enhance the SLMs’ ability to capture domain-specific contextual information and terminology distribution. 
Additionally, comparing baselines with ``Selector", ``Reasoner", and ``Selector+Reasoner (\textbf{S+R})'' confirms the effectiveness of the LLM Reasoner in enhancing reasoning capabilities. 
After incorporating the Reasoner on top of the Selector, the F1 scores are improved by 2.02\%, 0.96\%, and 1.92\% on the S\_Math, S\_Bio, and S\_Phy datasets, respectively.
These demonstrate that GRPO can effectively guide LLMs to align more closely with human annotation logic when reasoning about fine-grained concepts. 
Finally, integrating the Judger consistently enhances performance on all three datasets when comparing models using ``Selector+Reasoner+Judger (\textbf{S+R+J})" to those with ``Selector+Reasoner (\textbf{S+R})," highlighting the Judger's advantages in assessing the previous predictions from a global perspective. Thus, experimental results demonstrate that 
integrating all components is crucial, and a larger size of Judgers enhances the final performance of SRJudge. 


\begin{table}[t]
\centering
\small
\setlength{\tabcolsep}{1.75pt}
\renewcommand{\arraystretch}{1.0}
\begin{tabular}{l|cc|cc|cc}
\toprule
\multirow{2}{*}{\textbf{Position reward}} & \multicolumn{2}{c|}{\textbf{S\_Math}} & \multicolumn{2}{c|}{\textbf{S\_Bio}} & \multicolumn{2}{c}{\textbf{S\_Phy}} \\
 & ACC & F1 & ACC & F1 & ACC & F1 \\
\midrule
No   & 0.7559 & 0.7372 & 0.7397 & 0.6862 & 0.6806 & 0.6528 \\
\cline{1-7}
Static (Ours) & 0.7608 & 0.7451 & 0.7387 & 0.6888 & 0.6853 & 0.6619 \\
\cline{1-7}
Dynamic (Ours) & \textbf{0.7649} & \textbf{0.7491} & \textbf{0.7417} & \textbf{0.6925} & \textbf{0.6871} & \textbf{0.6629} \\
\bottomrule
\end{tabular}
\caption{Ablation study on position reward strategies in RL. Static position reward sets $R_{\text{position}}(o_p) = 3$, whereas Dynamic position reward follows the standard computation in Equation~\eqref{eq:rpos_and_alpha}.}
\label{tab:position}
\end{table}
Table \ref{tab:position} presents further ablation of the proposed dynamic position reward. 
SRJudge with position rewards consistently outperforms one without position rewards, which shows that introducing position rewards to the LLM Reasoner can prevent it from overfitting to the Selector’s outputs. This not only improves the tagging performance by encouraging exploration of other candidate positions but also accelerates model convergence.  
Furthermore, the comparison between these two rewards demonstrates that dynamically adjusting rewards across positions and training steps yields additional improvements. This is because, as training progresses, the model tends to become more conservative, and the dynamic reward helps sustain active exploration and improve RL efficiency. Additionally, to investigate the impact of the pruning rate, we evaluate annotation performance and training time across different pruning-rate settings in Stage 2 (RL), as shown in Table \ref{tab:pruning}. As the pruning rate increases, F1 scores exhibit a trend of initial improvement followed by a decline (while exceeding the no-pruning setting). Meanwhile, training time consistently decreases as the pruning rate increases. Overall, introducing the pruning mechanism not only improves macro F1 but also significantly enhances training efficiency. This is because, during the interaction between the agent and the environment, low-quality completions with positive but marginal advantages are effectively filtered out, reducing the agent’s exposure to noisy feedback and the number of gradient updates, and accelerating model convergence.
\begin{table}[t]
\centering
\setlength{\tabcolsep}{1pt}
\renewcommand{\arraystretch}{1.0}
\small
\begin{tabular}{c@{\hskip 1pt}|cc|cc|cc}
\toprule
 \multirow{2}{*}{\textbf{Pruning rate}}  & \multicolumn{2}{c|}{\textbf{S\_Math}} & \multicolumn{2}{c|}{\textbf{S\_Bio}} & \multicolumn{2}{c}{\textbf{S\_Phy}} \\
  & F1 &Time (h)   & F1 & Time (h)  & F1 & Time (h) \\
\midrule

0.00  		&0.7404 & 14.1 	&0.6821 & 10.4   &0.6603 &24.3\\
\cline{1-7}
0.25  		&0.7481 & 10.4 	&0.6912 & 6.7  &0.6612 &18.7 \\
\cline{1-7}
    0.50  		&\textbf{0.7491} & 7.2  		&\textbf{0.6925} & 5.9  &\textbf{0.6629} &13.4\\
\cline{1-7}
0.75  		&0.7408 & 5.5  		&0.6902 & 4.2  &0.6607 &9.1 \\
\bottomrule
\end{tabular}
\caption{Performance for RL using different pruning rates }
\label{tab:pruning}
\end{table}

\begin{figure}[t]
\centering
\includegraphics[width=0.475\textwidth]{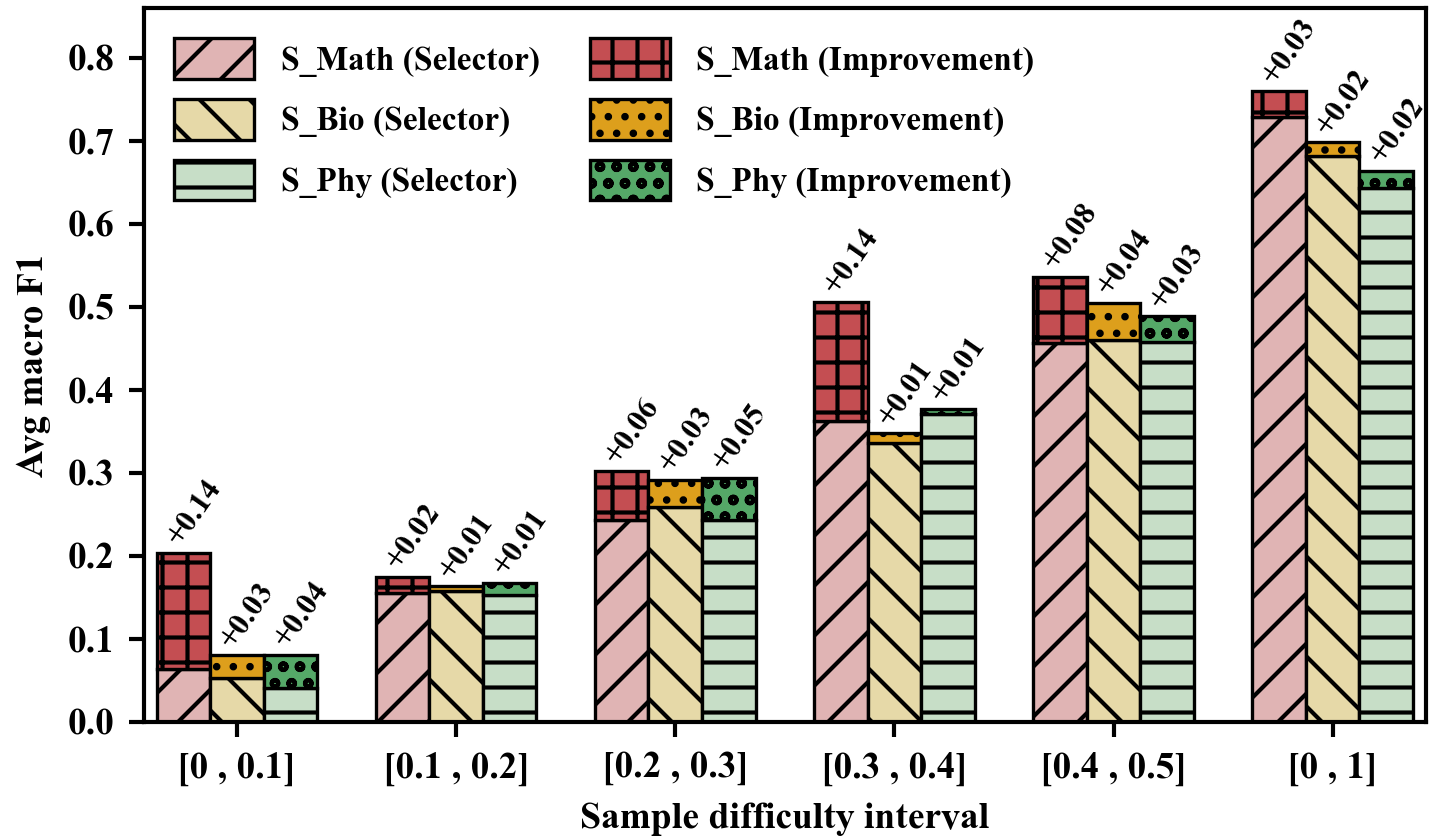} 
\caption{Performance of SRJudge on sample difficulty intervals.}
\label{fig:hard-to-distinguish}
\end{figure}

\subsection{Analysis on Hard-to-Distinguish Samples}
To evaluate the effectiveness of SRJudge on hard-to-distinguish samples, we use the Selector’s F1 score as a proxy for sample difficulty \cite{yang2025lgcel}, as illustrated in Figure \ref{fig:hard-to-distinguish}. Note that if the F1 score for the knowledge concept category predicted by the Selector falls below 0.5, that category is deemed ``difficult to distinguish". Using this criterion, we set five difficulty intervals as: [0, 0.1], [0.1, 0.2], [0.2, 0.3], [0.3, 0.4], and [0.4, 0.5].
For each difficulty interval, we evaluate the incremental effects brought by incorporating SRJudge. 
Experimental results show that 
SRJudge consistently achieves performance improvements across most difficulty intervals. These gains mainly arise from the enhanced reasoning and semantic understanding capabilities of the Reasoner trained via reinforcement learning, while the Judger further improves overall performance by integrating global contextual information and jointly evaluating the outputs of the Selector and the Reasoner. Notably, the performance gains of SRJudge on the higher-grade datasets S\_Bio and S\_Phy are relatively limited compared to those on S\_Math, primarily because the knowledge concepts in these disciplines are more complex and interrelated, which hinders accurate annotation. 
Overall, SRJudge achieves average F1 improvements of 8.84\% on S\_Math, 2.31\% on S\_Bio, and 3.24\% on S\_Phy, demonstrating the clear effectiveness of SRJudge in improving knowledge concept prediction for hard-to-distinguish samples across different disciplines.

\section{Conclusion}
This paper presents a novel three-stage SRJudge framework that enhances LLMs with selective reasoning via reinforcement learning, enabling selection from numerous candidates. This method effectively integrates the reasoning strengths of LLMs with the efficient selection capabilities of small language models. To the best of our knowledge, this is the first work to customize RL to improve LLM reasoning over fine-grained knowledge concepts. Additionally, we construct two high-quality datasets to support follow-up research herein. 
Experimental results across datasets demonstrate the effectiveness of SRJudge for knowledge concept tagging, offering a streamlined framework that empowers LLMs with selective reasoning to distinguish fine-grained knowledge concepts.


\appendix

\section*{Ethical Statement}

There are no ethical issues.

\section*{Acknowledgments}
This work is supported by the research project funded by Guangdong Basic and Applied Basic Research Foundation (2026A1515011829, 2024A1515140144), the Fundamental Research Funds for the Central Universities (21624325, 21624338), Ministry of Education of the People's Republic of China Humanities and Social Sciences Youth Foundation (24YJC890034), the Key Laboratory of Smart Education of Guangdong Higher Education Institute, Jinan University (2022LSYS003). This work is also partially supported by NSFC (62377028,62077028), Science and Technology Planning Project of Guangzhou Development District (2023GH01),”Master Mentor Plan” of Jinan University (YDXS2501), the Fundamental Research Funds for the Central Universities (21625102), the teaching reform research projects of Jinan University (JG2026030).

\bibliographystyle{named}
\bibliography{main}

\end{document}